\documentclass[10pt,twocolumn,letterpaper]{article}

\usepackage{iccv}              

\usepackage{booktabs}
\usepackage{multirow}
\usepackage{pifont}
\newcommand{\cmark}{\ding{51}}
\newcommand{\xmark}{\ding{55}}

\definecolor{iccvblue}{rgb}{0.21,0.49,0.74}
\usepackage[pagebackref,breaklinks,colorlinks,allcolors=iccvblue]{hyperref}

\def\paperID{5509} 
\def\confName{ICCV}
\def\confYear{2025}

\title{AnimateAnyMesh: A Feed-Forward 4D Foundation Model \\for Text-Driven Universal Mesh Animation}

\author{Zijie Wu$^{1,2}$\thanks{Work done during internship at DAMO Academy, Alibaba Group\\$\dagger$ Corresponding author}, Chaohui Yu$^{2,3}$, Fan Wang$^{2}$, Xiang Bai$^{1\dagger}$\\
\\
$^1$ Huazhong University of Science and Technology
$^2$ DAMO Academy, Alibaba Group
$^3$ Hupan Lab\\
{\tt\small \{zjw1031,xbai\}@hust.edu.cn, \{huakun.ych,fan.w\}@alibaba-inc.com}
\\
\url{https://animateanymesh.github.io/AnimateAnyMesh/}
}

\begin{document}

\makeatletter
\let\@oldmaketitle\@maketitle
\renewcommand{\@maketitle}{\@oldmaketitle
 \centering
    \includegraphics[width=1.0\linewidth]{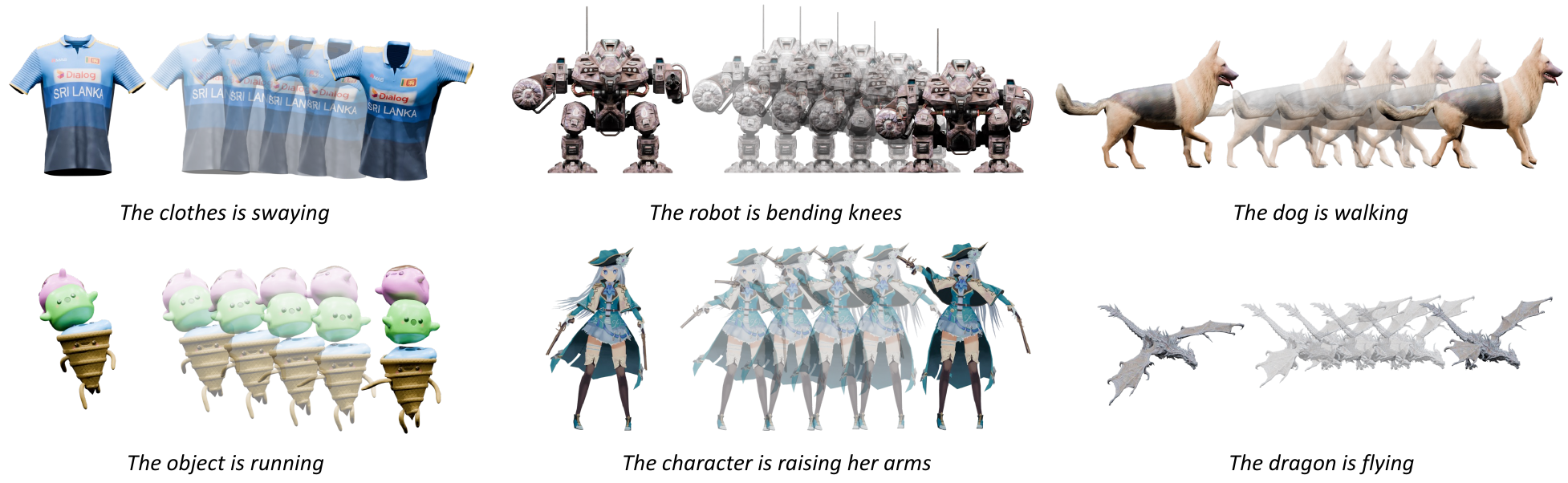}
    \vspace{-.15in}
    \captionof{figure}{We present \textbf{AnimateAnyMesh}: the first feed-forward universal mesh animation framework that enables efficient motion generation for arbitrary 3D meshes. Given a static mesh and prompt, our method generates high-quality animations in only \textbf{a few seconds}.}
    \label{fig:teaser}
  \bigskip}
\makeatother

\maketitle

\vspace{-.2in}

\begin{abstract}
Recent advances in 4D content generation have attracted increasing attention, yet creating high-quality animated 3D models remains challenging due to the complexity of modeling spatio-temporal distributions and the scarcity of 4D training data. In this paper, we present AnimateAnyMesh, the first feed-forward framework that enables efficient text-driven animation of arbitrary 3D meshes. Our approach leverages a novel DyMeshVAE architecture that effectively compresses and reconstructs dynamic mesh sequences by disentangling spatial and temporal features while preserving local topological structures. To enable high-quality text-conditional generation, we employ a Rectified Flow-based training strategy in the compressed latent space. Additionally, we contribute the DyMesh Dataset, containing over 4M diverse dynamic mesh sequences with text annotations. Experimental results demonstrate that our method generates semantically accurate and temporally coherent mesh animations in a few seconds, significantly outperforming existing approaches in both quality and efficiency. Our work marks a substantial step forward in making 4D content creation more accessible and practical. All the data, code, and models will be open-released.
\end{abstract}    
\section{Introduction}
\label{Sec:intro}

The revolution in 3D content creation has transformed various domains like VR/AR and gaming. While recent generative models~\cite{instant3d,lgm,12345,12345++,lrm,meshlrm,instantmesh,grm,gs-lrm,zhang2025lpm} excel at producing high-quality 3D assets, extending these advances to 4D content generation remains challenging due to the complexity of spatio-temporal modeling and scarcity of high-quality 4D assets.

Existing 4D generation approaches fall into two categories: per-scene optimization methods~\cite{sc4d,consistent4d,4dfy,animate124,dreamgaussian4d} and multi-view dynamic video generation methods~\cite{animate3d,4diffusion,diffusion4d}. The former employs SDS~\cite{dreamfusion} with pre-trained generative models but suffers from high computational costs and inconsistency. The latter 
fine-tunes video generation models~\cite{mvdream,animatediff,zeroscope,modelscope} on multi-view renderings of 4D assets~\cite{objaverse,objaverse-xl}, followed by per-scene 4D reconstruction. Although inference efficiency improves, the need for post-processing impedes real-time application. Moreover, these methods, which typically adopt dynamic 3DGS~\cite{3dgs} or NeRF~\cite{nerf} as 4D representations, suffer from view discrepancies due to the lack of ground-truth 4D data, relying solely on multi-view rendering supervision.

Given these limitations, we argue that dynamic meshes serve as an ideal representation for 4D content creation. As the de facto standard in modern graphics pipelines, mesh representations not only offer superior rendering efficiency but also enable natural decoupling of geometry and motion. Moreover, leveraging existing high-quality 3D meshes for animation, rather than pursuing direct 4D generation, enjoys several advantages: First, there exists a vast repository of high-quality 3D meshes, whether crafted by artists~\cite{sketchfab,objaverse} or synthesized by generative methods~\cite{meshlrm,instantmesh}, while high-quality 4D assets remain scarce. Second, decomposing 4D generation into geometry creation and motion modeling allows for higher fidelity and better control in both aspects. These observations motivate us to focus on the fundamental task of text-driven mesh animation.

To this end, we propose \textbf{AnimateAnyMesh}, the first feed-forward framework for text-driven universal mesh animation. At its core, we introduce \textbf{DyMeshVAE}, a novel VAE~\cite{vae} architecture tailored for dynamic mesh sequences, aiming to compress and reconstruct the trajectory of each vertex. Specifically, DyMeshVAE first decomposes each trajectory into the initial vertex position and the relative trajectory. These components are then encoded using distinct positional encoding mechanisms, mapping spatial and temporal features into high-dimensional spaces.
To enhance the quality of compression and reconstruction, we leverage the mesh topology by constructing the vertex connectivity matrix from the face information. This matrix serves as an attention mask to encode connectivity information into vertex features, which helps preserve topological structures and prevents trajectory entanglement during reconstruction (check Sec.~\ref{dymeshvae} for details). 
During decoding, the enhanced vertex features of the initial mesh serve as queries to retrieve corresponding relative trajectories through cross-attention, enabling accurate trajectory reconstruction.
Through this hierarchical design, DyMeshVAE effectively compresses meshes with varying numbers of vertices/faces into a fixed number of tokens (512 as default) during training while maintaining high-quality reconstruction. Moreover, being entirely attention-based, DyMeshVAE can dynamically adjust the number of encoded tokens during inference to accommodate meshes of different complexities, enabling animation of more intricate geometries.

To bridge the gap between text descriptions and mesh animations, we propose \textbf{Shape-Guided Text-to-Trajectory Model}, which leverages a Rectified Flow-based~\cite{rf} training strategy in the compressed latent space. By learning the conditional distribution of relative trajectories given text prompts and initial mesh features, our approach enables the generation of smooth and realistic animations. To facilitate this learning process, we introduce the \textbf{DyMesh Dataset}, a large-scale collection of over \textbf{4M} dynamic mesh sequences comprising about \textbf{2.6M} 16-frame and \textbf{1.6M} 32-frame sequences. This comprehensive dataset is carefully curated from diverse 4D assets through rigorous collection, filtering, processing, and annotation procedures, providing a robust foundation for training and evaluation.

Training with our DyMesh Dataset, AnimateAnyMesh achieves remarkable capabilities in generating high-quality animations that faithfully align with text descriptions for meshes of arbitrary topology. The feed-forward nature of our framework enables real-time inference, establishing AnimateAnyMesh as the first solution that combines universal mesh animation with both versatility and efficiency.

We highlight the contribution of this paper as follows:

\begin{enumerate}
    \item We propose AnimateAnyMesh, the first feed-forward framework for text-driven universal mesh animation, enabling generation of high-quality animations for meshes of arbitrary topology within a few seconds.
    \item We introduce DyMeshVAE, a novel compression-reconstruction architecture that effectively handles dynamic mesh sequences through trajectory decomposition and topology-aware attention mechanisms, naturally scaling to meshes of varying complexities.
    \item We curate DyMesh Dataset, a large-scale collection of over 4M dynamic mesh sequences with text annotations, providing a robust foundation for 4D generation.
    \item Extensive experiments demonstrate that our approach achieves state-of-the-art performance in text-driven mesh animation, combining high fidelity, versatility, and computational efficiency.
\end{enumerate}

\section{Related Works}

\begin{figure*}[t]
  \centering
   \includegraphics[width=0.96\linewidth]{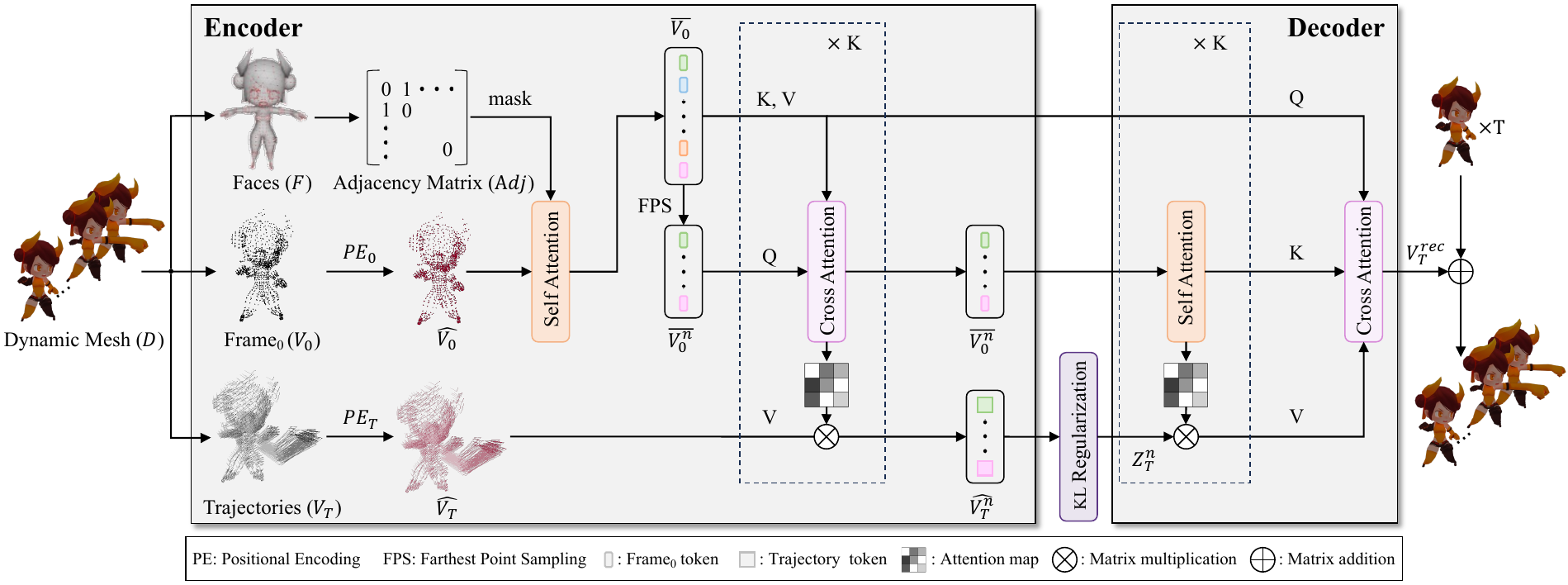}
   \vspace{-.1in}
   \caption{Illustration of our proposed \textbf{DyMeshVAE}. Given a dynamic mesh $D$, we first extract the initial frame vertex $V_0$, the connectivity information from faces $F$, and the relative trajectories $V_T$. These information are then encoded to a decoupled latent space $\{\overline{V_0^n}, \widehat{V_T^n}\}$ via the \textbf{Encoder}, which features trajectory decomposition and topology-aware attention mechanisms. Then the relative trajectories $V^{rec}_T$ are reconstructed from the latent space via the \textbf{Decoder}. Finally, we add $V^{rec}_T$ and $V_0$ to get the reconstructed dynamic mesh.}
   \label{fig:dvae}
   \vspace{-.1in}
\end{figure*}

\noindent\textbf{3D Generation.}
Early approaches~\cite{dreamfusion,dream3d,sjc,magic3d,prolificdreamer,dreamcraft3d,pointsto3d,make-it-3d,dreamgaussian,gsgen,gaussiandreamer} for 3D generation leverage CLIP~\cite{clip} score or Score Distillation Sampling (SDS)~\cite{dreamfusion} to distill geometric priors from pre-trained 2D generative models~\cite{sd, imagen}. However, due to the inherent lack of 3D information, these methods often suffer from view discrepancy issues and require time-consuming per-scene optimization, significantly limiting their practical applications.
To address these limitations, some methods~\cite{zero123,imagedream,syncdreamer,zero123++,richdreamer} try to fine-tune 2D generative models using multi-view renderings of 3D assets~\cite{objaverse,objaverse-xl}, thereby enhancing view consistency in the generated 3D content. Nevertheless, these approaches still require per-instance reconstruction after obtaining multi-view outputs.
In contrast, more recent methods~\cite{instant3d,lgm,12345,12345++,lrm,meshlrm,instantmesh,grm,gs-lrm,zhang2025lpm} directly generate 3D representations~\cite{3dgs,nerf} and optimize the networks through multi-view rendering supervision, enabling rapid 3D asset generation within seconds for given prompts. This end-to-end approach circumvents the need for expensive post-processing optimization.
Inspired by the evolution of 3D generation techniques, we pioneer a feed-forward 4D architecture for universal mesh animation. Our approach enables rapid mesh animation in a few seconds without per-scene optimization or reconstruction.


\noindent\textbf{4D Generation.}
Following the evolution path of 3D generation, early 4D generation approaches~\cite{sc4d,consistent4d,4dfy,animate124,dreamgaussian4d,alignyg,mav3d,4dgen,efficient4d,stag4d} attempt to distill spatio-temporal priors from 2D/3D/video generative models. Compared to 3D generation, these distillation approaches for 4D content creation not only demand significantly more computing power and longer optimization time to generate a single scene, but also tend to produce more noticeable spatio-temporal artifacts.
Some recent works~\cite{animate3d,4diffusion,diffusion4d,l4gm,vividzoo} have explored finetuning 3D/video generative models~\cite{mvdream,animatediff,zeroscope,modelscope} with 4D data to synthesize multi-view dynamic videos, aiming to accelerate 4D generation and improve spatio-temporal consistency. Nevertheless, these approaches still rely heavily on per-scene 4D reconstruction, and the generated objects exhibit spatial discrepancies due to the lack of true 4D training data. While some methods~\cite{motion2vecsets,humanmdm,motiondiffuse,realistichumanmt,physdiff} have achieved efficient 4D generation in specific categories (e.g., human bodies) through parametric models~\cite{smpl} and modality-specific data~\cite{dt4d,AMASS,AMASS_CMU,AMASS_ACCAD,AMASS_BMLhandball,AMASS_BMLmovi,AMASS_BMLrub,AMASS_DanceDB,AMASS_DFaust,AMASS_EyesJapanDataset,AMASS_GRAB,AMASS_GRAB-2,AMASS_HDM05,AMASS_HUMAN4D,AMASS_HumanEva,AMASS_KIT-CNRS-EKUT-WEIZMANN,AMASS_KIT-CNRS-EKUT-WEIZMANN-2,AMASS_KIT-CNRS-EKUT-WEIZMANN-3,AMASS_MoSh,AMASS_MOYO,AMASS_PosePrior,AMASS_SFU,AMASS_SOMA,AMASS_TCDHands,AMASS_TotalCapture,AMASS_WheelPoser}, to the best of our knowledge, direct feed-forward 4D generation for general categories remains unexplored.
In this work, we propose a novel framework that enables efficient feed-forward mesh animation during inference by disentangling and compressing dynamic mesh shapes and motions. Then, we leverage Rectified Flow~\cite{rf} to effectively model the posterior distribution of object motions conditioned on text and shape latents. This approach marks a significant advancement toward general-purpose feed-forward 4D content generation.

\section{Method}

Existing approaches to mesh animation typically rely on computationally intensive per-instance optimization~\cite{dreamgaussian4d,animate3d,motiondreamer} or are constrained to specific object categories~\cite{humanmdm,motiondiffuse,realistichumanmt,physdiff}, limiting their practical applications. To address these limitations, we propose AnimateAnyMesh, a feed-forward framework that enables text-driven universal mesh animation. Our framework consists of two key components: DyMeshVAE and Shape-Guided Text-to-Trajectory Model. The former effectively handles meshes of arbitrary topology by decomposing them into initial frames and relative trajectories, which are then compressed into a structured latent space, while the latter learns to generate trajectory features conditioned on both the initial mesh latent and text embeddings. In this section, we elaborate on the architecture of DyMeshVAE and Shape-Guided Text-to-Trajectory Model in Sec.~\ref{dymeshvae} and Sec.~\ref{rf_model}, respectively, followed by the training and inference pipeline of in Sec.~\ref{diff_pipe}.

\subsection{DyMeshVAE}
\label{dymeshvae}

\noindent\textbf{DyMeshVAE Encoder}.
As shown in Fig.~\ref{fig:dvae}, given a dynamic mesh sequence $D\subset \{F\in\mathbb{R}^{M\times3}, V\in\mathbb{R}^{T\times N\times 3}\}$, we first disentangle the vertices sequence $V$ into the initial frame vertices $V_0\in\mathbb{R}^{N\times 3}$ and the relative trajectories $V_T\in\mathbb{R}^{N\times (T\cdot 3)}$, which satisfy the following: 

\begin{equation}
  V^t = V_0^t + V_T^t, \quad \text{for } t = 1, 2, \dots, T
  \label{eq:v_sep}
\end{equation}
where $t$ stands for the index of the time sequence. 

We decompose the vertex sequence $V$ into initial positions $V_0$ and relative trajectories $V_T$ based on our empirical observation that such decomposition leads to the disentanglement of shape and motion, while yielding a motion distribution that better approximates a zero-mean normal distribution.
Following this decomposition, each trajectory is represented as a combination of its initial vertex position and the subsequent temporal offsets. The initial position serves as a spatial identifier, while the temporal offsets become our modeling target. To enhance trajectory reconstruction stability and prevent adhesion effects (as demonstrated in Fig.~\ref{fig:abl2}), we get inspirations from~\citep{3dshape2vecset,nerf} and employ distinct positional encoding schemes for $V_0$ and $V_T$, resulting in encoded features $\widehat{V_0}$ and $\widehat{V_T}$.

\begin{figure}[t]
  \centering
   \includegraphics[width=0.95\linewidth]{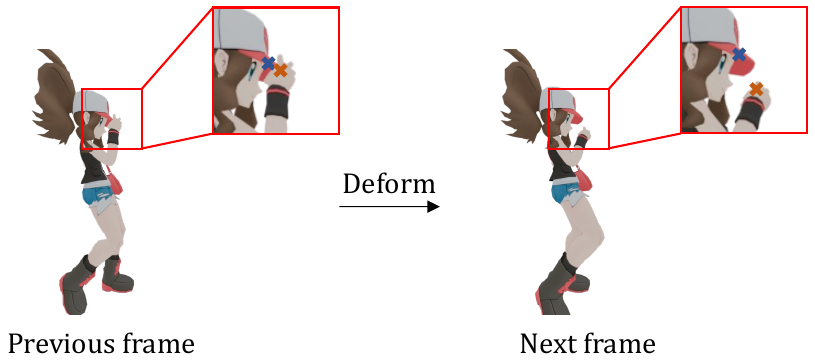}
   \vspace{-.1in}
   \caption{Demonstration of divergent trajectories for nearby mesh vertices in the initial frame.}
   \label{fig:traj_stuck}
   \vspace{-.1in}
\end{figure}

Due to the inherent local rigidity of objects, vertices that are spatially proximate in the initial frame typically exhibit similar motion trajectories. However, this spatial proximity assumption breaks down when considering vertices that belong to different structural components of the object, as illustrated in Fig.~\ref{fig:traj_stuck} (compare the trajectories of the \textcolor{blue}{blue} and \textcolor{orange}{orange} crosses). To address this limitation, we exploit the mesh's topological structure and design a topology-aware attention mechanism by incorporating connectivity information into vertex features, which serve as more reliable trajectory identifiers. Specifically, we first construct an adjacency matrix $Adj$ from the face information $F$ of the input mesh. This adjacency matrix then acts as an attention mask in a self-attention layer, enabling each vertex to aggregate information from its connected neighbors:

\begin{equation}
  \overline{V_0} = \text{Softmax}\left( \frac{\widehat{V_0}\cdot \widehat{V_0}^T \odot Adj}{\sqrt{d_k}  }\right) \widehat{V_0} + \widehat{V_0},
  \label{eq:v_aggre}
\end{equation}
where $d_k$ denotes the channel dimension of the projected space, and $\odot$ denotes Hadamard product.

Building upon our observation that vertices with similar neighborhood-enhanced features tend to exhibit similar motion patterns, we leverage this correlation by applying Farthest Point Sampling (FPS)~\cite{pointnet++} to the topology-aware vertex features $\overline{V_0}$ to obtain $\overline{V_0^n}$.The same sampling indices are used to sample $\widehat{V_T}$, resulting in $\widehat{V_T^n}$. 
Subsequently, we employ a cross-attention mechanism where $\overline{V_0^n}$ serves as the query, while the complete set of topology-aware features $\overline{V_0}$ acts as both key and value. The resulting attention map is then used to project $\widehat{V_T}$, as formulated below:

\begin{equation}
  \overline{V_0^n} = \text{Softmax}\left( \frac{\overline{V_0^n}\cdot \overline{V_0}^T}{\sqrt{d_k}  }\right) \overline{V_0} + \overline{V_0^n},
  \label{eq:enc_ca_v0}
\end{equation}
\begin{equation}
  \widehat{V_T^n} = \text{Softmax}\left( \frac{\overline{V_0^n}\cdot \overline{V_0}^T}{\sqrt{d_k}  }\right) \widehat{V_T} + \widehat{V_T^n}.
  \label{eq:enc_ca_vt}
\end{equation}

We employ a stack of cross-attention layers, as described above, and iteratively apply Equ.~(\ref{eq:enc_ca_v0}), (\ref{eq:enc_ca_vt}) to progressively gather global information into $\overline{V_0^n}$ and $\widehat{V_T^n}$.

\noindent\textbf{KL Regularization}.
After encoding, a dynamic mesh sequence is compressed to a pair of latents $\{\overline{V_0^n}, \widehat{V_T^n}\}$. Since we focus on the task of animating existing meshes, we only have to model the distribution of the relative trajectories $\widehat{V_T^n}$.
In accordance with the latent diffusion framework~\cite{sd,3dshape2vecset}, our model employs KL-regularization in the latent space for the purpose of modulating feature diversity. Specifically, we linearly project $\widehat{V_T^n}$ by two fully-connected layers to predict the mean $\mu_T^n$ and standard variance $\sigma_T^n$ of the distribution of $\widehat{V_T^n}$. Then, the VAE latent is sampled as:

\begin{equation}
  Z_T^n = \mu_T^n + \sigma_T^n \cdot \epsilon, 
  \label{eq:vae_samp}
\end{equation}
where $\epsilon \sim \mathcal{N}(0,1)$. The KL regularization is as follows:

\begin{equation}
  L_{kl} = \frac{1}{2n\cdot C}\sum_{i=1}^{n}\sum_{j=1}^{C}((\mu_T^{i,j})^2 + (\sigma_T^{i,j})^2 - log(\sigma_T^{i,j})^2), 
  \label{eq:kl_loss}
\end{equation}
where $C$ denotes the number of channels.

\noindent\textbf{DyMeshVAE Decoder}. 
Similarly, to further exploit the correlation between motion patterns of vertices with similar neighborhood-enhanced features, we perform self-attention on the sampled topology-aware features $\overline{V_0^n}$. The resulting attention map is then used to project $Z_T^n$, capturing the motion relationships among vertices with similar local structural characteristics, which is formulated as:

\begin{equation}
  \overline{V_0^n} = \text{Softmax}\left( \frac{\overline{V_0^n}\cdot \overline{V_0^n}^T}{\sqrt{d_k}  }\right) \overline{V_0^n} + \overline{V_0^n},
  \label{eq:dec_sa_v0}
\end{equation}
\begin{equation}
  Z_T^n = \text{Softmax}\left( \frac{\overline{V_0^n}\cdot \overline{V_0^n}^T}{\sqrt{d_k}  }\right) Z_T^n + Z_T^n.
  \label{eq:dec_sa_vt}
\end{equation}

To enhance the model capacity, we stack $K$ identical blocks in a cascaded manner, where each block comprises the aforementioned attention mechanism.
Next, we perform cross attention where the encoded vertex features of the initial mesh $\overline{V_0}$ serve as the query, while $\overline{V_0^n}$ and $Z_T^n$ act as the key and value respectively, which can be formulated as:

\begin{equation}
  V_T^{rec} = \text{Softmax}\left( \frac{\overline{V_0}\cdot \overline{V_0^n}^T}{\sqrt{d_k}  }\right) 
  Z_T^n.
  \label{eq:dec_ca}
\end{equation}

The reconstructed trajectories are obtained by projecting $V_T^{rec}$ through a fully connected layer to map the channel dimension to $T\times3$.
The reconstruction loss is the mean square error between $V_T^{rec}$ and $V_T$:

\begin{equation}
  L_{rec} = \frac{1}{N} \sum_{i=1}^{N} \|(V_T^{rec})^i - V_T^i\|_2^2.
  \label{eq:rec_loss}
\end{equation}

The overall loss is a weighted combination of $L_{rec}$ and $L_{kl}$: $L_{dvae}=L_{rec}+\gamma \cdot L_{kl}$, where we set $\gamma=0.001$ as default.
Finally, we transpose the obtained relative trajectories $V_T^{rec}$ and add the initial vertex positions $V_0$ to get the reconstructed dynamic mesh.

\subsection{Shape-Guided Text-to-Trajectory Model}
\label{rf_model}
\begin{figure}[t]
  \centering
   \includegraphics[width=1.0\linewidth]{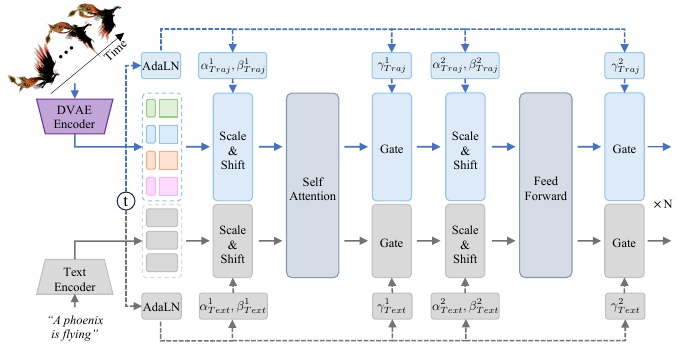}
   \vspace{-.2in}
   \caption{The architecture of the Shape-Guided Text-to-Trajectory Model. DVAE stands for the proposed DyMeshVAE.}
   \label{fig:rf}
   \vspace{-.2in}
\end{figure}

For text-driven mesh animation, our model learns to estimate the posterior distribution of relative trajectories conditioned on both the initial mesh and textual prompts.
As shown in Fig.~\ref{fig:rf}, we get inspirations from~\citep{sd3,cogvideox,dit} and propose the Shape-Guided Text-to-Trajectory Model, which builds upon the MMDiT~\cite{sd3} architecture. 

As described in Sec.~\ref{dymeshvae}, each dynamic mesh sequence $D$ is encoded into a pair of latent representations $\{\overline{V_0^n}, Z_T^n\}$. To facilitate effective multi-modal learning, we first normalize these latents using their respective global statistics (mean $\mu_0, \mu_T$ and standard deviation $\sigma_0, \sigma_T$) to eliminate numerical disparities. The normalized features are then concatenated along the channel dimension to form a comprehensive trajectory embedding. Meanwhile, text prompts are encoded through a pre-trained CLIP~\cite{clip} text encoder to obtain text embeddings.
Following~\citep{sd3,cogvideox}, we apply separate Adaptive Layer Normalization (AdaLN) parameters conditioned on timestep t to modulate the trajectory and text embeddings, respectively. These rescaled features are subsequently concatenated for self-attention computation. This modality-specific normalization strategy effectively bridges the distributional gap between varied representations, enabling more robust multi-modal learning. After processing through multiple such attention blocks, we decompose the output features back into and restore their original scales using the preserved statistics $\mu_0, \mu_T$ and $\sigma_0, \sigma_T$.

\subsection{Diffusion Pipeline}
\label{diff_pipe}
\noindent\textbf{Training.}
Following the diffusion paradigm in Rectified Flow (RF)~\cite{rf}, we aim to minimize the mean square error between the predicted and ground truth flow. To be noted, since $\overline{V_0^n}$ is settled, our model focuses solely on learning the distribution of the relative trajectory features $Z_T^n$. Specifically, we apply the diffusion process to $Z_T^n$ to obtain its noisy version $\tilde{Z_T^n}$, as formulated below:

\begin{equation}
  \tilde{Z_T^n}=(1-t)Z_T^n+t\epsilon,~~t = 1-\frac{1}{tan(\frac{\pi}{2}u)+1},
  \label{eq:rf}
\end{equation}
where $\epsilon\sim \mathcal{N}(0,1)$ and $u\sim \mathcal{U}(0,1)$.
Then, the optimization process can be formulated as:

\begin{equation}
  L_{rf} = \mathbb{E}_{\overline{V_0^n}, C_{text}}\left \|  v_\theta(\tilde{Z_T^n};\overline{V_0^n}, C_{text}) - u_T^n \right \| _2^2,
  \label{eq:rf_loss}
\end{equation}
where $v_\theta$ represents the backbone in Sec.~\ref{rf_model}. And $u_T^n=Z_T^n-\epsilon$, where $\epsilon$ is the sampled noise.

\noindent\textbf{Inference.}
We implement the sampling procedure for Rectified Flow following the flow-based ODE formulation. The sampling process can be formulated as:

\begin{equation}
  \frac{d\tilde{Z_T^n}}{dt} = f_\theta (\tilde{Z_T^n}(t),t), ~~Z_T^n(1)\sim \mathcal{N}(0,1) ,
  \label{eq:rf_samp}
\end{equation}
where $f_\theta$ denotes the learned velocity field.
For classifier-free guidance (CFG), we modify the velocity field as:

\begin{equation}
  f_\theta^{cfg} = f_\theta^{uncond} + \gamma \cdot(f_\theta^{cond}-f_\theta^{uncond}),
  \label{eq:cfg}
\end{equation}
where we set the scale $\gamma=3.0$ as default.

After sampling, we feed both the sampled features $Z_T^n$ and the topology-aware vertex features $\overline{V_0^n}$ into the decoder to generate the relative trajectories. The final animated mesh sequence is generated by applying the decoded vertex displacements to the vertices of the given mesh.
\section{Experiments}
\subsection{Settings}

\begin{figure*}[t]
  \centering
   \includegraphics[width=1.0\linewidth]{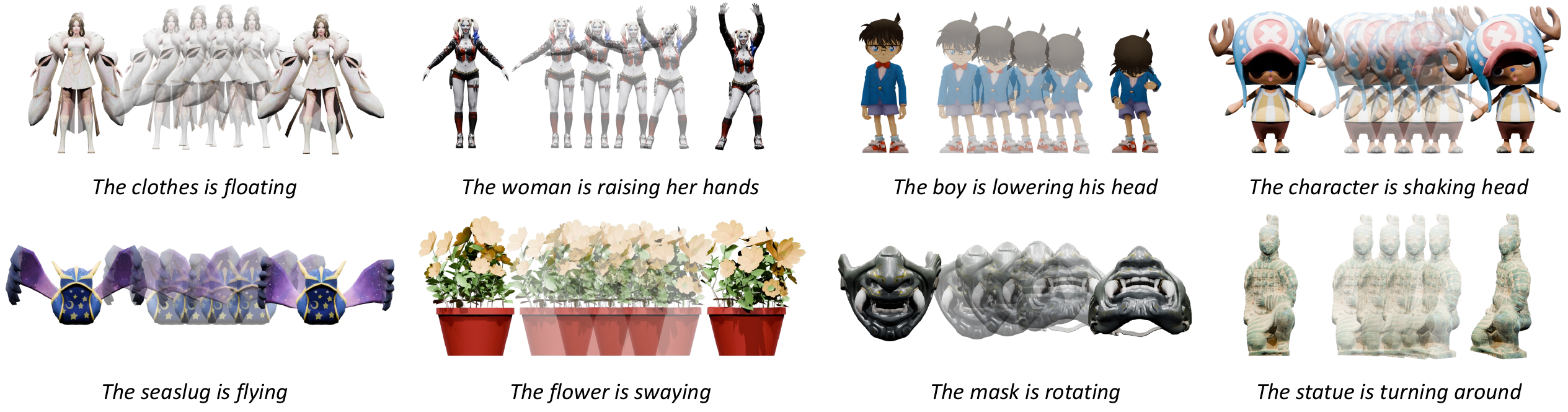}
   \vspace{-.3in}
   \caption{Animation examples of AnimateAnyMesh. Our model demonstrates the capability to generate high-quality and semantically plausible mesh animations for arbitrary input meshes based on text prompts. Best viewed when zoomed in.}
   \label{fig:results}
   \vspace{-.1in}
\end{figure*}


\noindent\textbf{Data Curation.}
As mentioned in Sec.~\ref{Sec:intro}, learning generalizable 4D generative models requires large-scale, high-quality 4D datasets. Recent works~\cite{animate3d,4diffusion,diffusion4d,l4gm,vividzoo} have attempted to construct such datasets by filtering and rendering multi-view video sequences from Objaverse~\cite{objaverse,objaverse-xl}. However, as discussed in Sec.~\ref{Sec:intro}, the lack of genuine 4D data in these datasets makes single-stage 4D generation impractical. To address this limitation, we convert various 4D assets into sequential dynamic meshes and propose DyMesh Dataset. Our dataset integrates multiple sources, including dynamic objects/scenes from Objaverse ($\sim$55k instances, counting multiple animations per object separately), dynamic human SMPL~\cite{smpl} models from AMASS~\cite{AMASS} ($\sim$8k instances), and dynamic human and animal models from DeformingThings4D~\cite{dt4d} ($\sim$2k instances), yielding a total of 66k complete animations. The final dataset statistics after sequence segmentation and filtering are presented in Tab.~\ref{tab:data}. For each sequence, we render front-view videos at $256\times 256$ resolution using Blender and generate captions using the Qwen-2.5-VL~\cite{qwen25} model. Detailed information about our dataset can be found in Sec.~\ref{data} of the appendix.

\begin{table}[t!]
  \centering
  \resizebox{\columnwidth}{!}{  
    \begin{tabular}{cc|ccc}
    \toprule
    \multirow{2}{*}{All Animations}             & \multirow{2}{*}{Frame number} & \multicolumn{3}{c}{Vertex number}   \\ \cline{3-5} 
                                                &                                    & $<=$50,000 & $<=$8192  & $<=$4096 \\ 
    \midrule
    \multicolumn{1}{c}{\multirow{2}{*}{66,209}} & \multicolumn{1}{c|}{16}             & 2,604,053  & 1,237,824 & 330,658  \\
    \multicolumn{1}{c}{}                        & \multicolumn{1}{c|}{32}             & 1,572,944  & 848,239   & 182,172  \\ \hline
    \end{tabular}
  }
  \vspace{-.1in}
  \caption{Statistical information of our DyMesh dataset.}
  \label{tab:data}
  \vspace{-.15in}
\end{table}

\begin{figure*}[t]
  \centering
   \includegraphics[width=1.0\linewidth]{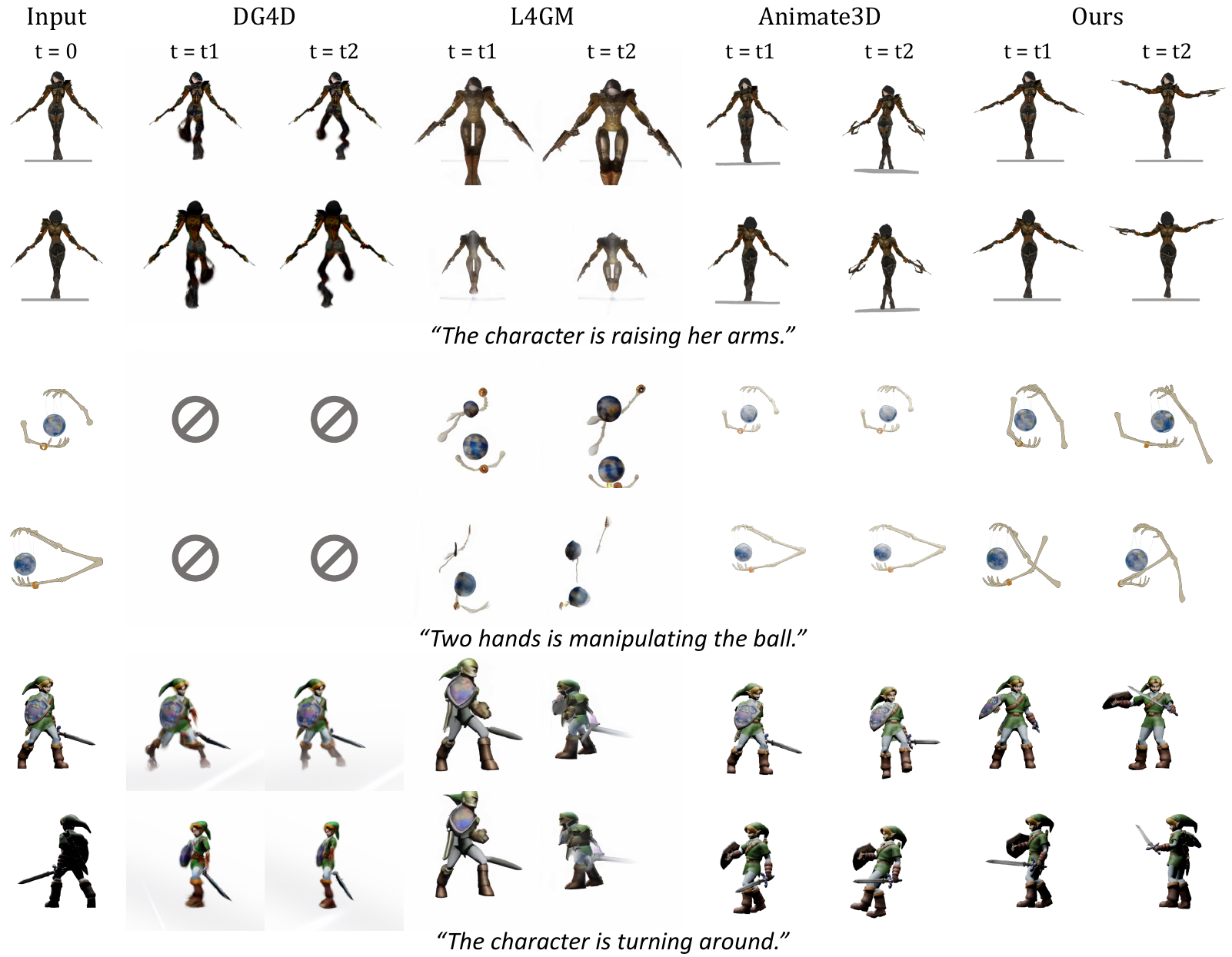}
   \vspace{-.25in}
   \caption{Qualitative comparison with state-of-the-art methods. $\oslash$ represents the failure cases. Best viewed when zoomed in.}
   \label{fig:qua}
   \vspace{-.1in}
\end{figure*}

\noindent\textbf{Implementation Details.}
We conduct extensive experiments to evaluate our proposed framework by training DyMeshVAE and the corresponding rectified flow model on four subsets of our dataset, varying the maximum vertex count (4096/8192) and sequence length (16/32 frames). For comparison with existing methods, we adopt the model variant trained on sequences with 4096 vertices and 16 frames. In this configuration, our DyMeshVAE architecture comprises an encoder with 8 cross-attention layers and a decoder with 8 self-attention layers, where the encoder samples 512 latent tokens via Farthest Point Sampling (FPS). The model is optimized using the Adam optimizer~\cite{adam} with a learning rate of 1e-4 for 1000 epochs on 8 H20 GPUs. 
For the rectified flow model, we leverage a pre-trained CLIP~\cite{clip} ViT-L/14 as the text encoder with a maximum sequence length of 77 tokens. The training is performed on 32 H20 GPUs using the Adam optimizer with a learning rate of 2e-4 for 1000 epochs.
At inference time, our framework demonstrates strong generalization capability by adapting to varying mesh resolutions through dynamic FPS sampling density adjustment. Specifically, processing a mesh with 20K vertices takes approximately 6 seconds on a single Nvidia A800 GPU for arbitrary mesh animation. We provide comprehensive implementation details in Sec.~\ref{imp} of the appendix.

\noindent\textbf{Evaluation Metrics.}
\label{metric}
To quantitatively evaluate DyMeshVAE's reconstruction quality, we compute the frame-wise average L2 distance between the reconstructed and ground-truth vertex trajectories. For evaluating mesh animations without references, we adopt the evaluation protocol adopted in Animate3D~\cite{animate3d}. Specifically, we render the animated meshes from specified viewpoints and leverage the view-aligned renderings as references to compute multiple perceptual metrics from VBench~\cite{vbench}, including I2V Subject Similarity, Motion Smoothness, and Aesthetic Quality (denoted as I2V, M.sm, and Aest.Q respectively). Note that we exclude the Dynamic Degree metric from VBench due to the presence of shape degeneration in some comparative methods under drastic motions.
To further assess the perceptual quality, we conduct a user study with 20 randomly recruited participants. Each participant is asked to rate the generated animations on a 5-point Likert scale (1-lowest, 5-highest) considering three aspects: text-to-motion alignment (User. Ta), motion naturalness (User. Mn), and shape preservation (User. Sp). The final scores are computed by averaging all ratings for each metric. Additionally, we report the computational efficiency by measuring the average time consumption for generating a single mesh animation. Comprehensive descriptions of evaluation metrics are provided in Sec.~\ref{supp_metric} of the appendix.

\subsection{Results}
Our proposed AnimateAnyMesh enables text-driven mesh animation for generic objects, generating dynamic mesh sequences that demonstrate sophisticated performance in semantic alignment, motion naturalness, and geometric consistency. Representative examples are illustrated in Fig.~\ref{fig:results}. Additional examples can be found in Fig.~\ref{fig:more_results} of the appendix.

\subsection{Comparison}
We evaluate our approach through comprehensive comparisons with three state-of-the-art methods, each employing distinct paradigms for multi-stage mesh animation. Specifically, we compare against DG4D~\cite{dreamgaussian4d}, L4GM~\cite{l4gm}, and Animate3D~\cite{animate3d}, all comparative evaluations are obtained using their official code and settings. We adopt DynamiCrafter~\cite{dynamicrafter} as the video generator for L4GM.

\noindent\textbf{Qualitative Comparison.}
As illustrated in Fig.~\ref{fig:qua}, existing methods exhibit various limitations in text-driven mesh animation. DG4D~\cite{dreamgaussian4d}, which distills knowledge from video diffusion models~\cite{svd} to optimize 3D Gaussians through SDS-based optimization, demonstrates limited shape preservation capability and suffers from severe object drifting, as evidenced in rows $3_{rd}$ and $4_{th}$ of Fig.~\ref{fig:qua}. L4GM~\cite{l4gm}, which aims to reconstruct 4D representations directly from single-view videos, is inherently limited by the performance of the video generator, particularly when handling rendered sequences of isolated 3D objects without natural backgrounds. While Animate3D~\cite{animate3d} proposes a theoretically sound approach by converting initial meshes to Gaussian Splatting representations and optimizing motion trajectories through multi-view dynamic video generation with ARAP~\cite{arap} regularization, its multi-stage pipeline suffers from error accumulation, leading to degraded performance in complex scenarios.
In contrast, our method achieves superior results through direct vertex trajectory prediction, demonstrating two key advantages: (1). efficient mesh animation through a feed-forward architecture, and (2). enhanced preservation of local geometric details while generating prompt-aligned realistic motions.
Please refer to Sec.~\ref{supp_qua} of the appendix for more visualization results.

\noindent\textbf{Quantitative Comparison.}
For quantitative comparison, we curate a test set comprising 10 randomly selected objects across multiple categories (human, animals, weapons, etc.). Using identical text prompts as conditions, we compute the mentioned metrics (detailed in Sec.~\ref{metric}) across all baseline methods. The comparative results are presented in Tab.~\ref{tab:quant}.

       

\begin{table}[t!]
  \centering
  \resizebox{\columnwidth}{!}{  
    \begin{tabular}{l|ccc|ccc|l}
      \toprule
      \multirow{2}{*}{Method}                           & \multicolumn{3}{c|}{VBench}                                          & \multicolumn{3}{c|}{User study}                                              & \multicolumn{1}{c}{\multirow{2}{*}{Time~$\downarrow$}} \\ \cline{2-7}
                                                  & I2V~$\uparrow$ & M.Sm~$\uparrow$ & Aest.Q~$\uparrow$ & User.Ta~$\uparrow$ & User.Mn~$\uparrow$ & User.Sp~$\uparrow$ & \multicolumn{1}{c}{}                                        \\
    \midrule
DG4D~\cite{dreamgaussian4d} & 0.811               & 0.926                & 0.476                  & 2.130                   & 2.460                   & 2.755                   & $\sim$10min                                                 \\
L4GM~\cite{l4gm}            & 0.844               & 0.992                & 0.464                  & 2.885                   & 2.865                   & 2.835                   & $\sim$30s                                                   \\
Animate3D~\cite{animate3d}  & 0.936               & 0.992                & 0.526                  & 2.850                   & 3.195                   & 3.405                   & $\sim$14min                                                 \\
Ours                                              & \textbf{0.954}      & \textbf{0.995}       & \textbf{0.539}         & \textbf{4.505}          & \textbf{4.700}          & \textbf{4.790}          & \textbf{$\sim$6s}                                           \\
      \bottomrule
    \end{tabular}
  }
  \vspace{-.1in}
  \caption{Quantitative comparison with state-of-the-art methods. All the evaluation is performed on a Nvidia A800 GPU.}
  \label{tab:quant}
  \vspace{-.1in}
\end{table}

As shown in Tab.~\ref{tab:quant}, our approach achieves superior performance across all VBench metrics (I2V, M.Sm, Aest.Q), indicating its effectiveness in both shape preservation and temporal motion coherence. The user study results further validate our method's advantages, showing substantial improvements in both text-motion alignment accuracy and motion naturalness. These quantitative evaluations align well with our qualitative observations. Moreover, AnimateAnyMesh significantly reduces the computational overhead during inference compared to existing methods, making it particularly promising for practical applications.

\subsection{Ablation Studies}
We conduct comprehensive ablation studies to validate the effectiveness of key components in our DyMeshVAE architecture. Specifically, we examine the incorporation of mesh adjacency information ($Adj$), positional encoding for both initial frame vertex features ($PE_0$) and relative trajectory features ($PE_T$), separate attention computation using initial frame features instead of complete trajectory features ($Sep~Attn$), and Farthest Point Sampling on neighborhood-enhanced vertex features ($\mathit{Emb~FPS}$) . Further ablation studies are provided in Sec.~\ref{supp_abl} of the appendix.

\begin{table}[t!]
  \centering
  \resizebox{0.8\columnwidth}{!}{  
    \begin{tabular}{ccccc|c}
      \toprule
      $Adj$ & $PE_0$ & $PE_t$ & $Sep~Attn$ & $\mathit{Emb~FPS}$ & Rec~Error $\downarrow$ \\
      \midrule
      \xmark & \cmark & \cmark & \cmark & \cmark & 0.500 \\
      \cmark & \xmark & \cmark & \cmark & \cmark & 0.443 \\
      \cmark & \cmark & \xmark & \cmark & \cmark & 0.441 \\
      \cmark & \cmark & \cmark & \xmark & \cmark & 0.478 \\
      \cmark & \cmark & \cmark & \cmark & \xmark & 0.291 \\
      \cmark & \cmark & \cmark & \cmark & \cmark & \textbf{0.223} \\
       
      \bottomrule
    \end{tabular}
  }
  \vspace{-.1in}
  \caption{Ablation studies of technical components of DyMeshVAE. Rec Error denotes the average L2 sum error per-instance.}
  \label{tab:abl}
  \vspace{-.1in}
\end{table}

\begin{figure}[t]
  \centering
   \includegraphics[width=1.0\linewidth]{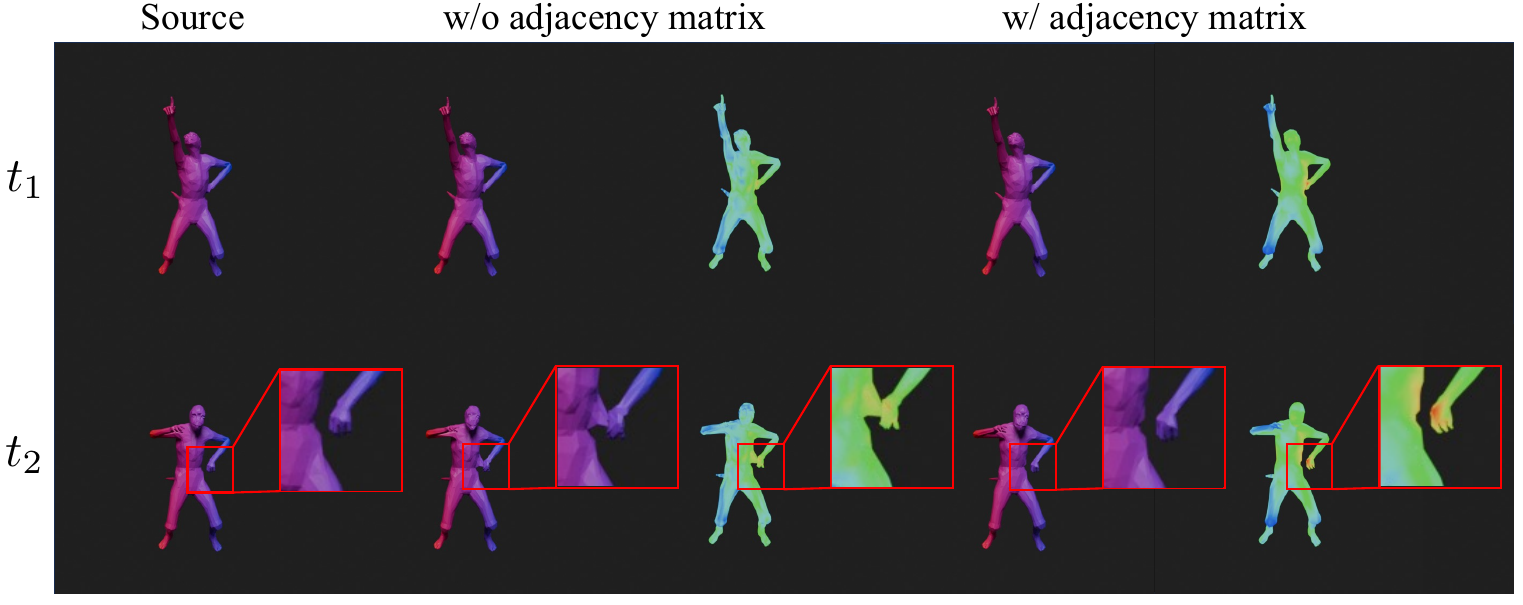}
   \vspace{-.25in}
   \caption{Ablation study on mesh adjacency information. Visualization shows the cosine similarity between feature embeddings of a query vertex (in hand region) and all mesh vertices, with similarity intensity ranging from \textcolor{blue}{blue} (low) to \textcolor{red}{red} (high).}
   \label{fig:abl1}
   \vspace{-.1in}
\end{figure}

\begin{figure}[t]
  \centering
   \includegraphics[width=1.0\linewidth]{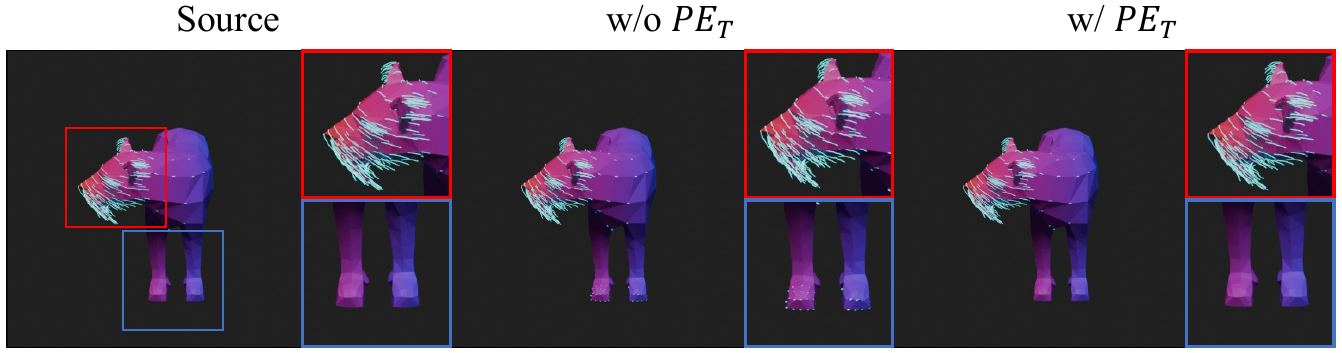}
   \vspace{-.25in}
   \caption{Ablation study on trajectory positional encoding ($PE_T$). Vertex trajectories shown in \textcolor{teal}{teal}. Please zoom in for a better view.}
   \label{fig:abl2}
   \vspace{-.15in}
\end{figure}

As shown in Tab.~\ref{tab:abl}, each architectural component in our framework plays a crucial role in ensuring high-quality mesh animation. The incorporation of mesh connectivity information ($Adj$) is essential for distinguishing vertices from different semantic regions. Without this topology information, vertices that are spatially adjacent but semantically distinct (e.g., hand and waist regions in Fig.~\ref{fig:abl1}) tend to be encoded similarly, leading to undesired adhesion artifacts during decoding. The mesh topology enables better feature discrimination through distinct neighborhood patterns, resulting in more accurate trajectory predictions.
Positional encoding schemes for both initial frame features ($PE_0$) and trajectory features ($PE_T$) project the low-dimensional trajectory information into a higher-dimensional space. This enhancement not only improves the discriminative power of queries but also ensures robust trajectory reconstruction, even under VAE sampling and noise perturbation. As demonstrated in Fig.~\ref{fig:abl2}, the absence of trajectory feature positional encoding leads to unstable motion patterns.
Furthermore, our design choice of computing attention maps using only initial frame features ($Sep~Attn$) effectively leverages the prior knowledge that vertices with similar neighborhood patterns tend to share similar trajectories. The implementation of Farthest Point Sampling ($\mathit{Emb~FPS}$) on topology-enhanced initial vertex features, rather than raw vertex coordinates, proves more effective in selecting representative trajectories, thereby enabling better compression and reconstruction capabilities.

\section{Conclusion}
\label{Sec:Con}
In this paper, we introduce AnimateAnyMesh, the first feed-forward framework for universal text-driven mesh animation. At its core lies DyMeshVAE, a novel VAE architecture specifically designed for dynamic mesh sequences. By decomposing dynamic meshes into initial states and relative trajectories while incorporating mesh topology information, DyMeshVAE achieves efficient variable-length compression and reconstruction of dynamic mesh sequences through carefully designed attention mechanisms.
To enable modeling the motion distribution in the compressed latent space, we propose the Shape-Guided Text-to-Trajectory Model, which adopts a Rectified Flow-based training strategy that effectively learns the distribution of trajectories conditioned on text and initial mesh features. To facilitate the training of our framework, we construct DyMesh Dataset, a large-scale dynamic mesh dataset containing over 4M dynamic mesh sequences.
Extensive experiments demonstrate that AnimateAnyMesh can generate high-quality language-driven animations for arbitrary meshes within seconds, marking a significant advancement in feed-forward 4D generation. 
{
    \small
    \bibliographystyle{ieeenat_fullname}
    \bibliography{main}
}

\appendix
\section{Settings}
In this section, we elaborate on the technical details of AnimateAnyMesh. The chapter is structured into three main components: data curation~\ref{data},  implementation details~\ref{imp}, and evaluation metric~\ref{metric}.
\subsection{Data Curation}
\label{data}
As described in the main paper, our curated 4D data is sourced from three primary sources: Objaverse~\cite{sketchfab,objaverse}, AMASS~\cite{AMASS}, and DT4D~\cite{dt4d}. Initially, we filter and extract all \texttt{.glb} files containing animation sequences from Objaverse. Using Blender's Python API (bpy), we convert each animation into a mesh sequence. Animations with fewer than 16 frames are discarded, and each sequence is capped at 200 frames. Post-conversion, each animation is encoded into a \texttt{.bin} file comprising 
$D\subset \{F\in\mathbb{R}^{M\times3}, V\in\mathbb{R}^{T\times N\times 3}\}$, where $M$ denotes the face count, $T$ represents the temporal length of the dynamic mesh sequence, and $N$ indicates the vertex count. Similarly, we develop scripts to convert SMPL~\cite{smpl} models from AMASS and \texttt{.anime} files from DT4D into the identical \texttt{.bin} format. 

Subsequently, we traverse all stored files, implementing vertex merging operations for duplicate vertices while updating the corresponding face information. This serves two purposes: data optimization and, crucially, supporting DyMeshVAE's encoding process, which embeds vertex connectivity information to prevent trajectory inconsistencies during decoding.

The processed dynamic mesh files undergo temporal slicing with window sizes $T=16$ and $T=32$. To maximize data utilization, we initiate slicing from both frame 0 and $T//2$, storing T-frame segments sequentially. We also preserve reverse-ordered sequences as independent files, effectively augmenting the dataset by 3-4$\times$. Each new sequence undergoes center normalization, positioning the initial frame at the origin with maximum vertex absolute values normalized to 1.0.

Post-slicing, we implement motion-based filtering, eliminating sequences with inter-frame maximum absolute differences outside the range [0.01, 0.5]. We also filter out the instances whose faces/vertices ratio exceeds 2.5. The cleaned data is then rendered using bpy scripts to generate frontal video sequences. We apply uniform gradient coloring (purple-red) and consistent top-down point lighting, utilizing the Cycles engine for $256\times 256$ resolution rendering on CPU clusters.

For caption generation, we employ Qwen-2.5-VL~\cite{qwen25} as our annotation model with the prompt: ``Describe the motion of the object in a sentence." The generated captions are stored alongside their corresponding \texttt{.bin} files.

Finally, we validate all processed files, removing examples with anomalous vertex or face shapes. The resulting DyMesh dataset is partitioned into subsets based on maximum vertex counts (4,096/8,192/50,000) to facilitate training and testing across different configurations.

\begin{figure*}[t]
  \centering
   \includegraphics[width=1.0\linewidth]{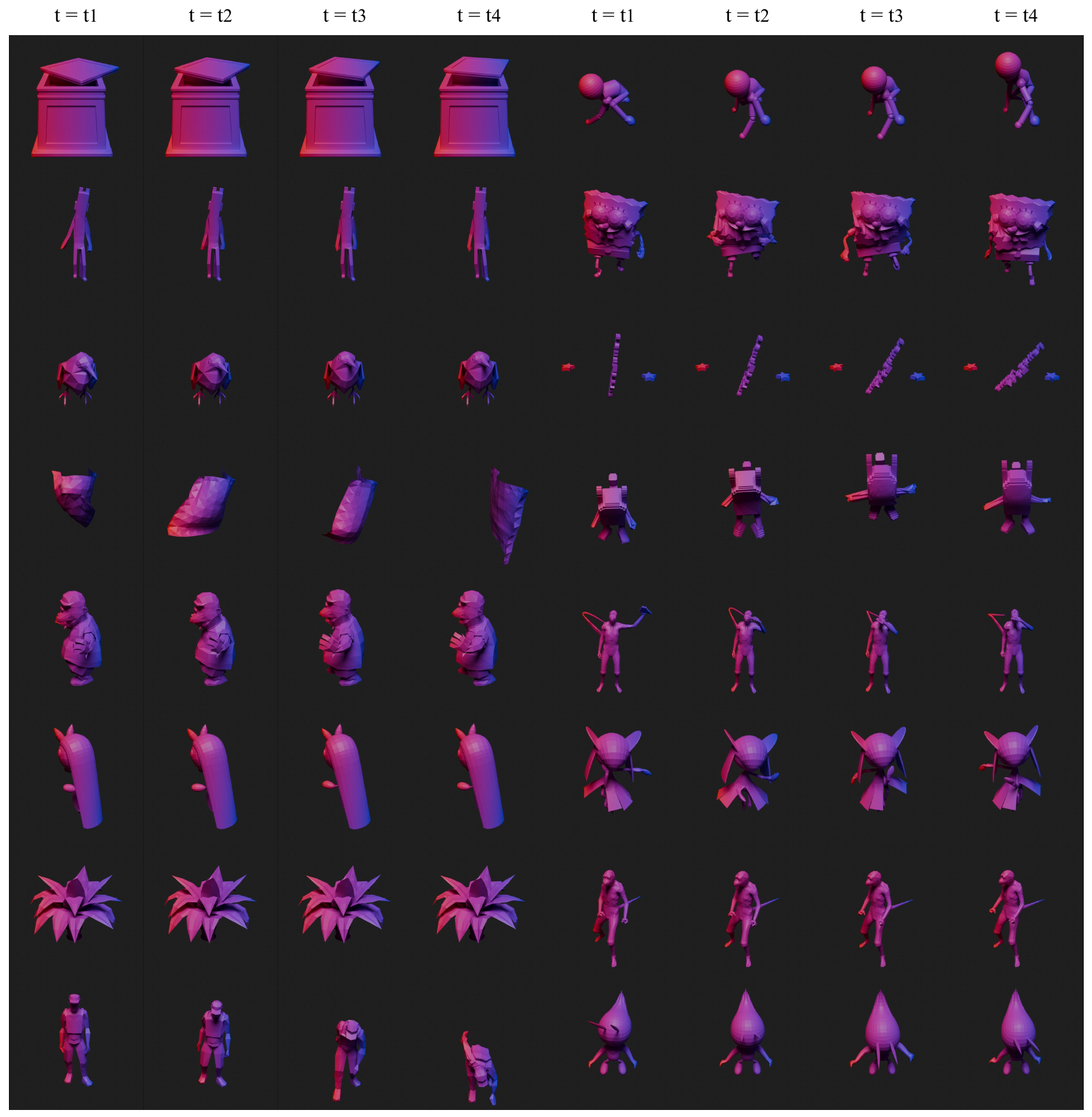}
   \caption{Examples of dynamic mesh sequences in our DyMesh dataset.}
   \label{fig:data}
\end{figure*}

\subsection{Implementation Details}
\label{imp}
In the DyMeshVAE architecture, both encoder and decoder utilize attention mechanisms with a hidden dimension of 512. For temporal settings of T=16 and T=32, we employ latent dimensions of 32 and 64 channels respectively for VAE sampling, with both configurations containing approximately 25M parameters. The Shape-Guided Text-to-Trajectory Model consists of 12 stacked transformer blocks as shown in Fig.~4 of the main paper, where each block incorporates 8-head attention layers with features projected to 512 dimensions, totaling approximately 200M parameters. We also conduct scaling experiments with an enhanced architecture of 740M parameters, comprising 24 transformer blocks, 16 attention heads per layer, and a 1024-dimensional latent space. During training, we implement an efficient batchify strategy where each sample's vertices and faces are padded to maintain uniform tensor dimensions across the batch: vertex tensors are padded with zero vectors (0.0, 0.0, 0.0) up to the dataset's maximum vertex count, while face indices are padded with invalid indices (-1, -1, -1) to reach the maximum face count (defined as 2.5 times the maximum vertex count), enabling consistent batch processing while preserving mesh topology integrity. During inference, we perform rectified flow sampling using 64 uniformly sampled timesteps within the [0,1] interval.

\subsection{Evaluation Metrics}
\label{supp_metric}
For quantitative evaluation, we employ three standardized metrics from VBench~\cite{vbench} to assess the performance of comparative methods: I2V Subject Consistency, Motion Smoothness, and Aesthetic Quality. Specifically, I2V Subject Consistency is computed by measuring the frame-wise similarity of DINO~\cite{dino} features, quantifying the visual coherence between the generated video and the reference image. Motion Smoothness is evaluated through the AMT~\cite{amt} video interpolation framework, which assesses the temporal continuity and fluidity of the generated motion sequences. The Aesthetic Quality metric leverages the LAION aesthetic predictor to quantify the perceptual quality and artistic value of individual frames from a human-centric perspective. These complementary metrics provide a comprehensive evaluation of both temporal consistency and visual fidelity of the generated results.

For our user study protocol, we recruit 20 participants from diverse backgrounds and age groups to evaluate comparative methods through a controlled assessment. We randomly selected 10 diverse test cases and generated motion sequences for each using all comparative methods based on text prompts, rendering each result from four orthogonal viewpoints and concatenating them temporally ($16\times 4=64$ frames). The results from all methods were randomly shuffled and horizontally concatenated into 64-frame GIFs, with participants rating each result on a 5-point Likert scale (5: excellent, 1: poor) across three criteria: text-motion alignment, motion plausibility, and shape preservation fidelity. The final evaluation scores were computed by aggregating and de-shuffling ratings across all participants, with failed generations being handled by computing means from successful cases only, ensuring a comprehensive and unbiased assessment of perceptual quality and semantic accuracy. All the User Study source videos can be found in the $User\_Study$ folder of the supplementary materials.

\begin{table}[t!]
  \centering
  \resizebox{0.7\columnwidth}{!}{  
    \begin{tabular}{ccccc}
      \toprule
      $num\_v$ & 4,096 & 8,192 & 16,384 & 32,768 \\
      \midrule
      t (s) & 3.95 & 5.99 & 10.68 & 21.86  \\
       
      \bottomrule
    \end{tabular}
  }
  \caption{Inference time evaluation. $num\_v$ represents the number of mesh vertices. We sample 1/8 number of vertices in the FPS sampling procedure as default. All these testing is conducted on a single Nvidia A800 GPU.}
  \label{tab:effi}
\end{table}

The inference efficiency of our framework scales with both mesh complexity (vertex/face count) and FPS feature sampling density. Our empirical studies indicate that an 8:1$\sim$4:1 ratio between vertex count and FPS samples achieves optimal performance-efficiency trade-off. The corresponding inference times across different mesh resolutions under this sampling configuration are presented in Tab.~\ref{tab:effi}.

\section{Animation Results of AnymateAnyMesh}
We curated a diverse collection of high-fidelity static meshes from Sketchfab~\cite{sketchfab}, encompassing various categories including humanoid figures, animals, weapons, clothes, and environmental assets. These meshes were animated using our proposed AnimateAnyMesh framework through text-driven synthesis. Fig.~\ref{fig:more_results} demonstrates representative results, showcasing our framework's capability to generate high-fidelity, versatile animations across arbitrary mesh topologies. The qualitative results validate the effectiveness of our approach in achieving generalized mesh animation with exceptional geometric fidelity, motion naturalness, and semantic flexibility.

Moreover, given identical input prompts and initial mesh as condition, our AnimateAnyMesh framework demonstrates robust multi-modal synthesis capabilities through different random seeds, generating diverse yet plausible high-fidelity mesh animations. Fig.~\ref{fig:seed} illustrates this generative flexibility through exemplar results, highlighting our framework's ability to explore varied motion manifestations while maintaining semantic consistency and geometric integrity.

\section{Additional Qualitative Comparison}
\label{supp_qua}
For a comprehensive comparison of mesh animation approaches, we present additional comparative examples against baseline methods in Fig.~\ref{fig:supp_qua}. The results consistently support our main findings, demonstrating that our approach outperforms existing methods in terms of text-motion alignment, motion naturalness, and shape preservation fidelity.

\section{Additional Ablation Studies}
\label{supp_abl}

\begin{figure}[ht]
  \centering
   \includegraphics[width=1.0\linewidth]{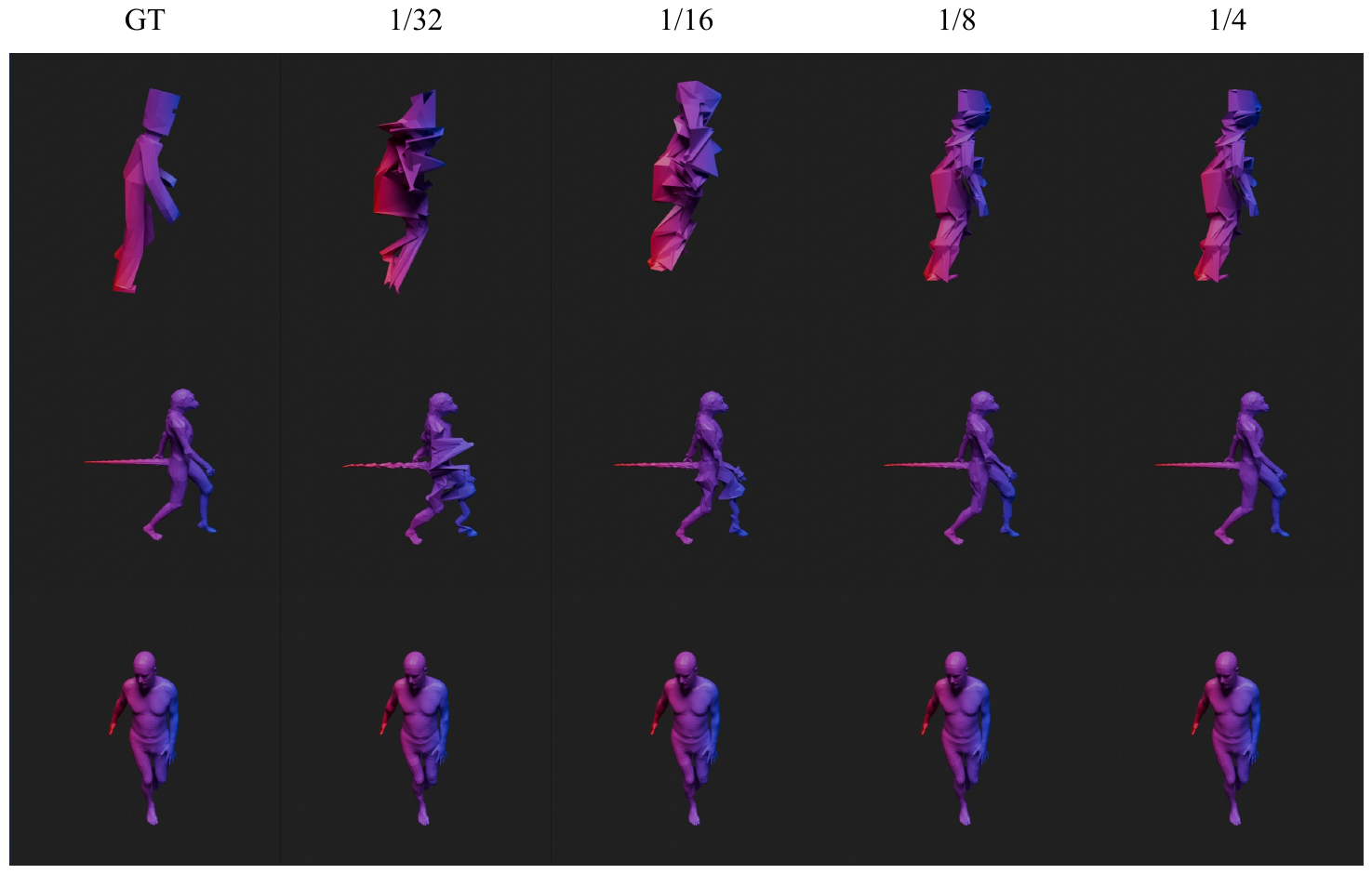}

   \caption{Ablation study on FPS Sampling Ratio. The numerical annotations above each image indicate the FPS ratio employed in the DyMeshVAE Encoder. Please zoom in for a better view.}
   \label{fig:samp_ratio}
\end{figure}

\begin{table}[ht]
  \centering
  \resizebox{0.7\columnwidth}{!}{  
    \begin{tabular}{c|cccc}
    \toprule
    \multirow{2}{*}{$num\_v$} & \multicolumn{4}{c}{FPS Sampling Ratio}   \\ \cline{2-5} 
                                                &                                     1/32 & 1/16 & 1/8 & 1/4 \\ 
    \midrule
     369 & 136.57 & 58.78 & 22.45 & 8.10  \\
     2,567 & 17.74 & 3.64 & 1.16 & 0.54  \\
     6,890 & 1.57 & 0.66 & 0.51 & 0.50  \\ \hline
    \end{tabular}
  }
  \caption{Reconstruction error for dynamic mesh sequences with varying vertex counts under different FPS sampling ratios.}
  \label{tab:samp_ratio}
\end{table}

\noindent\textbf{FPS Sampling Ratio.}
We conduct ablation study on the sampling ratio in the DyMeshVAE encoding procedure to see how it influences the reconstruction quality. We randomly sampled three dynamic mesh sequences from the DyMesh dataset with vertex counts of 369, ~2,567, and 6,890 (top to bottom). For each mesh, we conducted experiments using FPS (Farthest Point Sampling) ratios of 1/32, 1/16, 1/8, and 1/4 for feature sampling and reconstruction. The qualitative visualizations and quantitative metrics are presented in Fig.~\ref{fig:samp_ratio} and Tab.~\ref{tab:samp_ratio}, respectively, demonstrating the impact of sampling density on reconstruction fidelity.

As demonstrated in Fig.~\ref{fig:samp_ratio}, meshes with low vertex counts (row $1_{st}$) exhibit poor reconstruction quality even with a 1/4 sampling ratio, while high-vertex-count meshes (row $3_{rd}$) maintain satisfactory reconstruction fidelity even at a 1/32 sampling ratio, as validated by quantitative metrics in Table 3. We attribute this phenomenon to spatial distribution characteristics: low-vertex meshes typically exhibit sparse spatial distribution, leading to significant geometric information loss in regions surrounding unsampled vertices, whereas high-vertex-count meshes maintain dense surface coverage even with lower sampling ratios, enabling better preservation of local geometric features during encoding. Based on these observations, we set the feature sampling count to 512 during training to facilitate efficient batch processing while achieving an optimal balance between performance and computational efficiency across meshes with diverse vertex counts. During inference, we adopt an adaptive sampling strategy where n = min(512, $num\_v//8$) for inference, where $num\_v$ represents the number of mesh vertices. We empirically find the robust performance across the majority of test cases with this setting.

\section{Scaling Experiments}
In this section, we conduct scaling experiments across three dimensions: dataset scale, temporal resolution (frame count), and model capacity. We evaluate four configurations, each denoted as $\mathrm{rf\_<1>v\_<2>f\_<3>p}$, where $1,2,3$ represents the maximum number of vertices of the dataset, number of frames per instance, and the number of parameters of the Shape-Guided Text-to-Trajectory Model. All models were trained for 600,000 iterations on corresponding DyMesh subsets with a batch size of 2048 and a learning rate of 2e-4. For evaluation, we generate mesh animations using the same 10 mesh-prompt pairs from our qualitative benchmark, maintaining consistent random seeds across all models. Front-view renderings were produced to compute the VBench~\cite{vbench} metrics (I2V, M.Sm, Aest.Q) discussed in the main text. Additionally, given that all four trained models demonstrate the capability to generate high-quality and semantically plausible mesh animations, we incorporate the Dynamic Degree metric (abbreviated as Dy.Dg) from VBench to quantitatively assess motion intensity. The comprehensive results are presented in Tab.~\ref{tab:scale}.
The results indicate that: 
(1) Increasing the maximum number of vertices leads to better performance on most metrics (B vs. A).
(2) Increasing the number of frames will improve the output dynamic, improving Dy.Dg while maintaining promising results on other metrics. (C vs. A).
(3) Scaling the model's parameter size leads to better performance on all metrics, demonstrating good scalability of our method. (D vs. A).

\begin{table}[t!]
  \centering
  \resizebox{\columnwidth}{!}{  
    \begin{tabular}{c|ccccc}
      \toprule
      & Experiment & I2V~$\uparrow$ & M.Sm~$\uparrow$ & Aest.Q~$\uparrow$ & Dy.Dg~$\uparrow$ \\
    \midrule
A & $\mathrm{rf\_4096v\_16f\_200Mp}$ & 0.954      & 0.995       & 0.539       & 0.693  \\
B & $\mathrm{rf\_8192v\_16f\_200Mp}$ & \textbf{0.985}      & 0.996       & \textbf{0.550}       & 0.605  \\
C & $\mathrm{rf\_4096v\_32f\_200Mp}$ & 0.948      & 0.993       & 0.532       & \textbf{0.737}  \\
D & $\mathrm{rf\_4096v\_16f\_740Mp}$ & 0.968      & \textbf{0.997}       & 0.545       & 0.705  \\
      \bottomrule
    \end{tabular}
  }
  \caption{Scaling experiments of AnymateAnyMesh. We name the experiments as $\mathrm{rf\_<1>v\_<2>f\_<3>p}$, where $1,2,3$ represents the maximum number of vertices of the dataset, number of frames per instance, and the number of parameters of the Shape-Guided Text-to-Trajectory Model. }
  \label{tab:scale}
\end{table}

\section{Limitation}
Our work exhibits three limitations: First, regarding dataset scale, while the proposed DyMesh Dataset encompasses over 4M dynamic mesh sequences, the number of unique mesh identities remains below 100k, potentially limiting model generalization across specialized categories. We plan to address this by creating and curating additional high-quality, diverse 4D datasets. Second, concerning annotation quality, we observe that current video captioning models demonstrate suboptimal performance when annotating 3D rendered sequences without natural backgrounds, compared to their performance on natural videos, particularly in motion description granularity. Enhancing caption fidelity for synthetic 3D content remains a key research direction. Third, in terms of model capabilities, the current implementation of AnimateAnyMesh is confined to 16/32-frame sequence generation, and extending the model's capability to generate longer-duration mesh animations represents a significant future research objective.

\begin{figure*}[t!]
  \centering
   \includegraphics[width=1.0\linewidth]{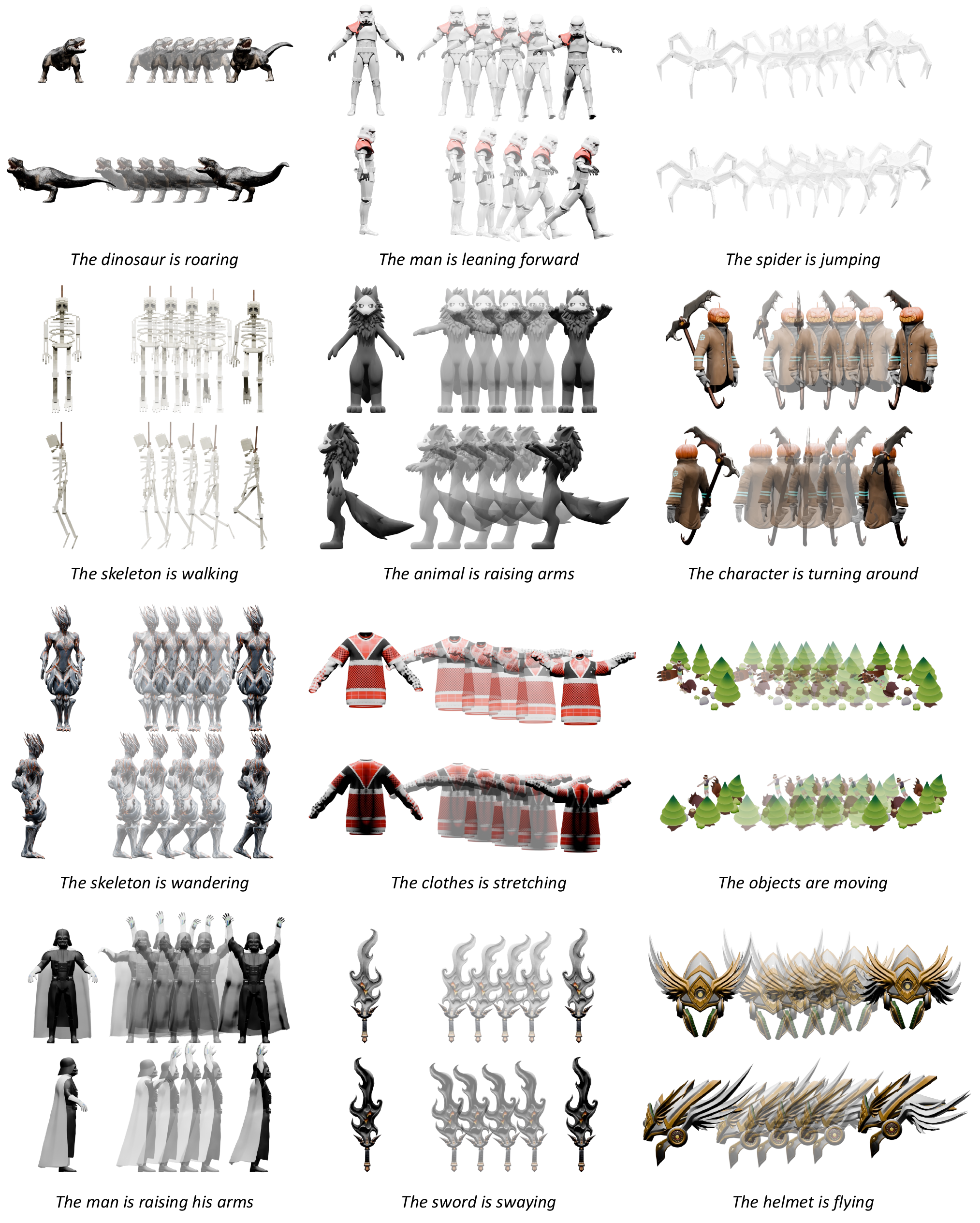}
   \caption{Examples of text-driven mesh animation results of the proposed AnimateAnyMesh. We render two random views for each example. Please zoom in for a better view.}
   \label{fig:more_results}
\end{figure*}

\begin{figure*}[t!]
  \centering
   \includegraphics[width=1.0\linewidth]{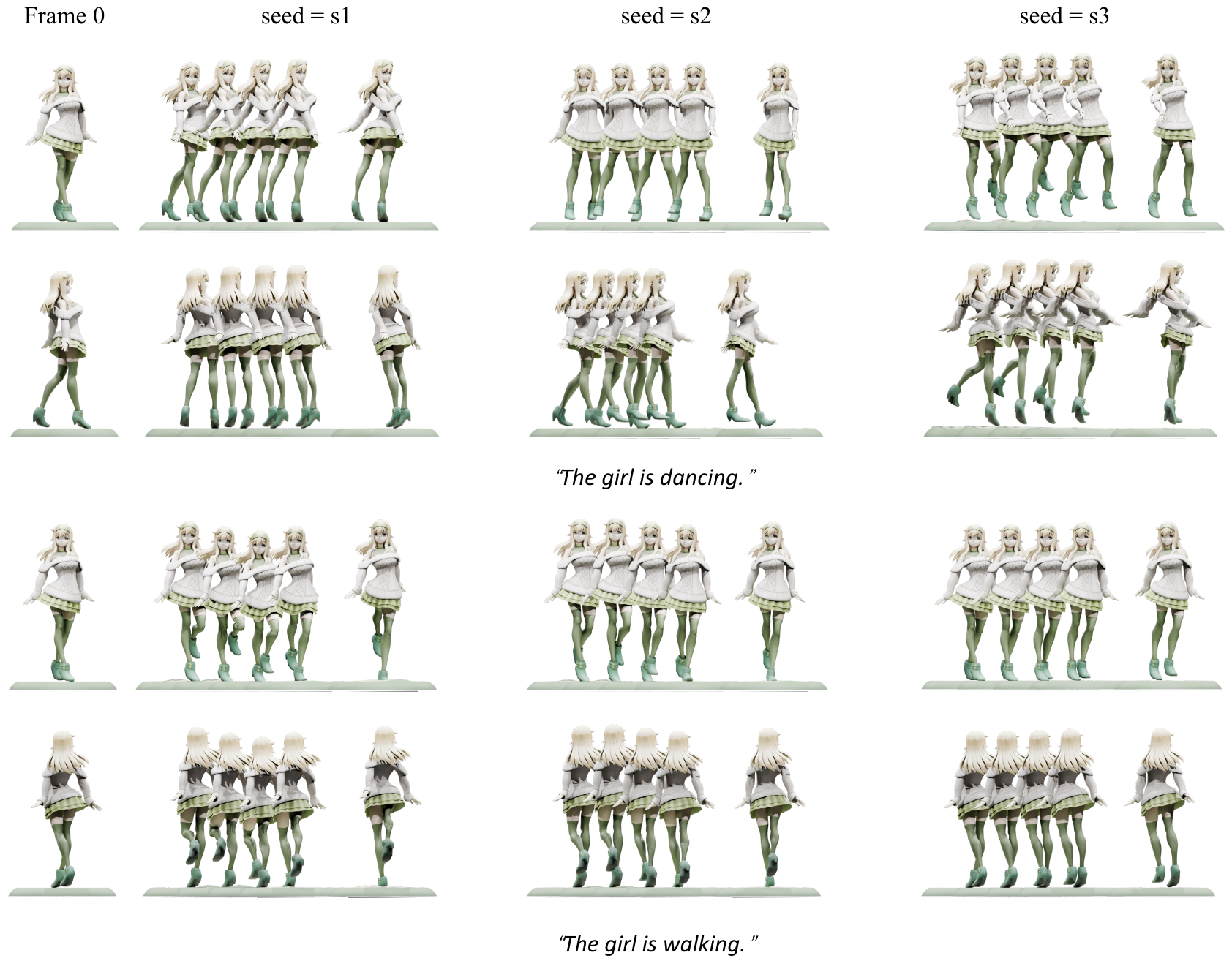}
   \caption{Diversity Demonstration of AnimateAnyMesh Generations. Given identical text-prompt and initial mesh conditions, AnimateAnyMesh demonstrates the capability to generate diverse, high-quality mesh animations through random seed variation.}
   \label{fig:seed}
\end{figure*}

\begin{figure*}[t!]
  \centering
   \includegraphics[width=1.0\linewidth]{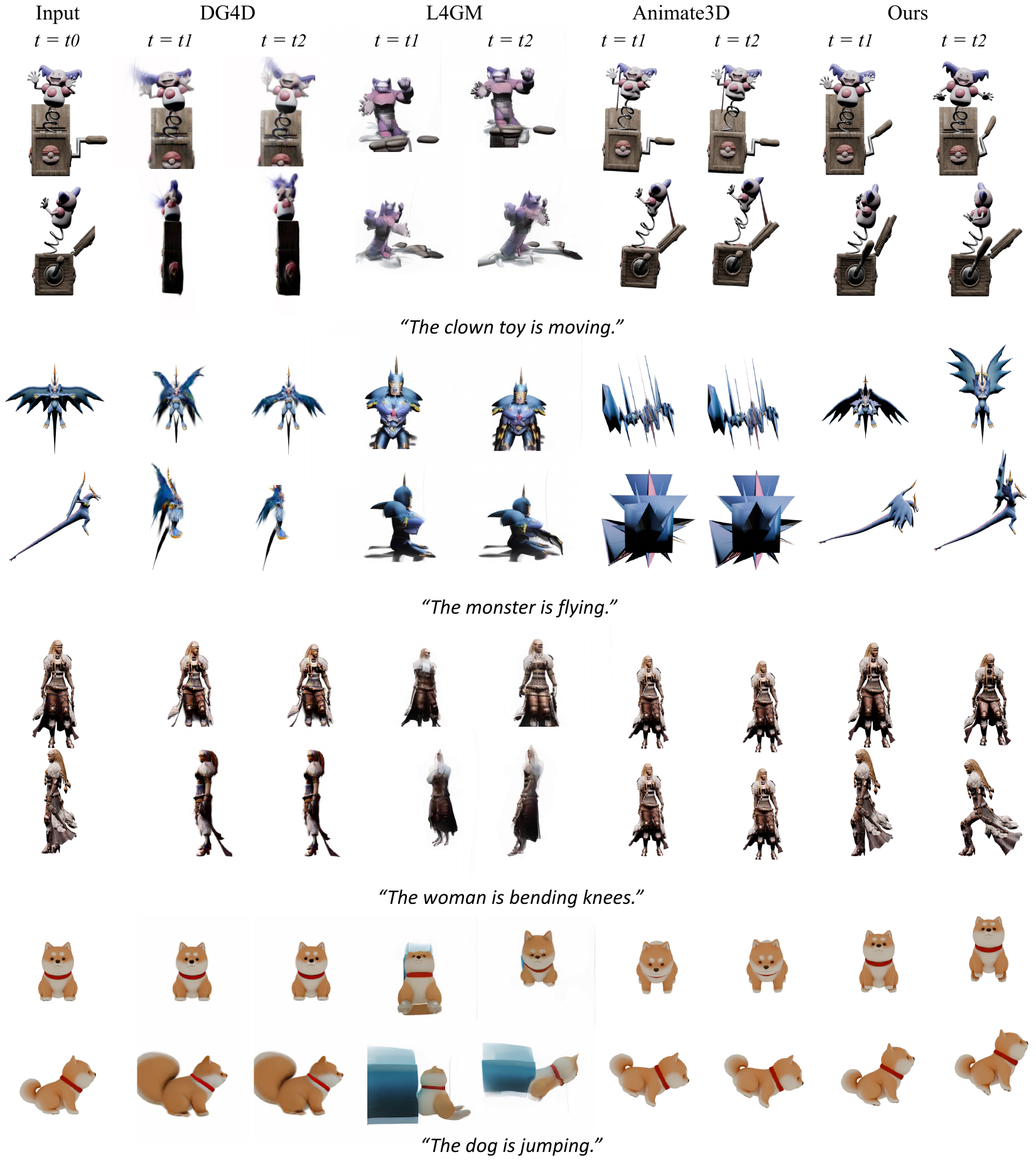}
   \caption{Additional qualitative comparison with state-of-the-art mesh animation methods.}
   \label{fig:supp_qua}
\end{figure*}

\end{document}